\documentclass[sigconf,nonacm]{acmart}

\usepackage[utf8]{inputenc}
\usepackage[T1]{fontenc}
\usepackage{booktabs}
\usepackage{amsmath}
\usepackage{graphicx}
\usepackage{subcaption}
\usepackage{placeins}
\usepackage{multirow}
\usepackage{array}
\usepackage{xcolor}
\usepackage{microtype}
\usepackage{enumitem}
\usepackage{capt-of}

\graphicspath{{figures/}{media/}}

\title{When Spike Sparsity Does Not Translate to Deployed Cost: VS-WNO on Jetson Orin Nano}

\author{Jason Yoo}
\email{jjyoo3@illinois.edu}
\orcid{0009-0004-5122-5477}
\affiliation{%
 \institution{University of Illinois Urbana-Champaign}
 \department{Department of Nuclear, Plasma \& Radiological Engineering}
 \city{Urbana}
 \state{IL}
 \country{USA}
}

\author{Shailesh Garg}
\email{Shailesh.Garg@am.iitd.ac.in}
\affiliation{%
 \institution{Indian Institute of Technology Delhi}
 \department{Department of Applied Mechanics}
 \city{New Delhi}
 \country{India}
}

\author{Souvik Chakraborty}
\email{souvik@am.iitd.ac.in}
\affiliation{%
  \institution{Indian Institute of Technology Delhi}
  \city{New Delhi}
  \country{India}
}
\affiliation{%
  \institution{University of Illinois Urbana-Champaign}
  \department{Department of Nuclear, Plasma \& Radiological Engineering}
  \city{Urbana}
  \state{IL}
  \country{USA}
}

\author{Syed Bahauddin Alam}
\email{alams@illinois.edu}
\affiliation{%
 \institution{University of Illinois Urbana-Champaign}
 \department{Department of Nuclear, Plasma \& Radiological Engineering}
 \city{Urbana}
 \state{IL}
 \country{USA}
}
\affiliation{%
 \institution{National Center for Supercomputing Applications}
 \city{Urbana}
 \state{IL}
 \country{USA}
}

\begin{document}

\begin{abstract}
Spiking neural operators are appealing for neuromorphic edge computing because event-driven substrates can, in principle, translate sparse activity into lower latency and energy. Whether that advantage survives deployment on commodity edge-GPU software stacks, however, remains unclear. We study this question on a Jetson Orin Nano 8\,GB using five pretrained variable-spiking wavelet neural operator (VS-WNO) checkpoints and five matched dense wavelet neural operator (WNO) checkpoints on the Darcy rectangular benchmark. On a reference-aligned path, VS-WNO exhibits substantial algorithmic sparsity, with mean spike rates decreasing from 54.26\% at the first spiking layer to 18.15\% at the fourth. On a deployment-style request path, however, this sparsity does not reduce deployed cost: VS-WNO reaches 59.6\,ms latency and 228.0\,mJ dynamic energy per inference, whereas dense WNO reaches 53.2\,ms and 180.7\,mJ, while also achieving slightly lower reference-path error (1.77\% versus 1.81\%). Nsight Systems indicates that the request path remains launch-dominated and dense rather than sparsity-aware: for VS-WNO, \texttt{cudaLaunchKernel} accounts for 81.6\% of CUDA API time within the latency window, and dense convolution kernels account for 53.8\% of GPU kernel time; dense WNO shows the same pattern. On this Jetson-class GPU stack, spike sparsity is measurable but does not reduce deployed cost because the runtime does not suppress dense work as spike activity decreases.
\end{abstract}

\ccsdesc[500]{General and reference~Performance}
\ccsdesc[300]{Computer systems organization~Embedded and cyber-physical systems}
\ccsdesc[300]{Computing methodologies~Neural networks}

\keywords{neural operators, spiking neural networks, edge deployment, Jetson Orin Nano, Darcy flow, neuromorphic}

\maketitle

\section{Introduction}
Neural operators are attractive for edge scientific computing because they can replace expensive PDE solves with learned surrogates at inference time. For such deployment settings, however, model quality alone is not sufficient; the relevant question is whether the learned operator delivers useful latency, energy, and memory behavior on the target platform. This question is especially important for spiking variants, whose efficiency claims rely on sparse activity but whose deployed cost is ultimately determined by the execution stack.

VS-WNO introduces spiking dynamics into wavelet neural operators and reports substantial spike sparsity on PDE benchmarks~\cite{garg2024vswno}. That sparsity lowers deployed cost only when the execution stack reduces work with spike activity. Commodity edge-GPU stacks need not satisfy this condition because dense kernels, dense memory traffic, and fixed launch overhead may remain. We therefore ask whether the sparsity of VS-WNO changes deployed latency and energy on an actual edge device.

We evaluate five pretrained VS-WNO checkpoints and five matched dense WNO checkpoints on the Darcy rectangular benchmark using a Jetson Orin Nano 8\,GB. We separate a \emph{reference-aligned} path for error and layer-wise spike activity from a \emph{deployment-style} path for request-serving cost, including preprocessing, inference, decoding, board-level energy, and system RAM. Across five seeds, VS-WNO exhibits substantial spike sparsity yet remains slower and higher-energy than the matched dense baseline on the measured request path. Nsight Systems traces show a launch-dominated, dense execution path rather than a sparsity-aware one. Our contributions are:
\begin{itemize}[nosep,leftmargin=*]
\item We measure substantial layer-wise spike sparsity in VS-WNO, but find no latency or energy benefit on Jetson Orin Nano.
\item We compare against matched dense WNO checkpoints and find that dense WNO is faster, lower-energy, and slightly more accurate on the same deployment path.
\item We use Nsight Systems to show why sparsity is not converted into deployed efficiency on the evaluated stack: the request path remains launch-dominated and dense.
\item We show that on Jetson's unified-memory architecture, system RAM is a more deployment-relevant capacity metric than CUDA-reserved memory.
\end{itemize}

\section{Background}

\subsection{VS-WNO}
Wavelet neural operators (WNOs) parameterize solution operators in wavelet space by inserting discrete wavelet transforms (DWTs) and inverse transforms inside each recursive update block~\cite{tripura2023wavelet}. VS-WNO replaces the GeLU activations in those blocks with variable spiking neuron (VSN) layers implemented in snnTorch~\cite{garg2024vswno,eshraghian2023training}. Each VSN maintains a leaky membrane potential with learnable decay~$\beta$ and threshold parameters and emits a graded spike when the threshold is crossed. Training jointly minimizes a PDE loss and a spike-rate regularizer, encouraging sparse activity while preserving predictive accuracy. On the Darcy rectangular benchmark, the original work reports $1.81 \pm 0.02\%$ relative $L^2$ error~\cite{garg2024vswno}.

\subsection{Deployment context}
Algorithmic sparsity and deployed efficiency are not the same quantity. On neuromorphic hardware, inactive neurons need not trigger the same work as active ones because the execution model is event-driven~\cite{davies2018loihi}. On a conventional PyTorch/CUDA stack, the runtime instead dispatches dense cuDNN convolutions, matrix operations, and elementwise kernels, often with fixed launch and memory costs independent of activation sparsity. Unless the stack introduces explicit sparsity-aware execution paths, sparse activations alone do not imply reduced executed work. This distinction is especially relevant on embedded GPUs, where unified memory, software overheads, and launch behavior can dominate arithmetic cost. While recent neural-operator work has begun to emphasize edge deployability and hardware portability for real-time virtual sensing on irregular domains~\cite{howes2026graphneuraloperatoredge}, the present question is narrower. We ask whether model-side sparsity on a commodity edge-GPU stack is actually converted into reduced executed work and lower deployed cost.

\section{Experimental Setup}

\subsection{Platform}
All experiments run on a Jetson Orin Nano 8\,GB edge-GPU platform (Table~\ref{tab:platform}) in MAXN SUPER mode with \texttt{jetson\_clocks} enabled.

\begin{table}[t]
\centering
\caption{Jetson Orin Nano 8\,GB platform configuration.}
\label{tab:platform}
\setlength{\tabcolsep}{4pt}
\begin{tabular}{@{}ll@{}}
\toprule
GPU & Ampere, 1024 CUDA cores, 32 Tensor Cores \\
Memory & 8\,GB LPDDR5 unified (CPU\,+\,GPU shared) \\
Peak BW & 102\,GB/s (MAXN\_SUPER) \\
CUDA / cuDNN & 12.6.85 / 9.20.0.48 \\
OS / L4T & Ubuntu 22.04 / L4T 36.4.7 \\
Mode & MAXN\_SUPER, \texttt{jetson\_clocks} \\
\bottomrule
\end{tabular}
\end{table}

\subsection{Workload}
We study the Darcy rectangular benchmark used in the original VS-WNO work: a 2-D Darcy flow equation on a rectangular domain, where the model learns the mapping from permeability $a(x,y)$ to pressure $u(x,y)$ on an $85 \times 85$ grid~\cite{garg2024vswno}. We evaluate two pretrained model families provided through collaboration: five VS-WNO checkpoints and five matched dense WNO checkpoints, each with seeds $\{0,10,20,30,50\}$. The VS-WNO checkpoints use the db4 wavelets, four recursive update layers, width~64, direct coordinate inputs, trainable VSN parameters, and a single spike time step. The dense WNO checkpoints share the same wavelet operator structure, width, and input/output setup, but replace spiking nonlinearities with standard dense activations.

For reference-aligned evaluation, we follow the original data split and compute normalization statistics from 1000 training samples, then evaluate all 100 test samples. For deployment-style evaluation, we do not reload the full benchmark files at inference time. Instead, we form a compact request workload from the same test set using an 8-sample request bundle together with precomputed normalization statistics. This setup reflects a realistic serving scenario in which model weights and normalization parameters are already resident in memory.

\subsection{Measurement harness}
We maintain two execution paths. The \emph{reference-aligned} path loads the full training and test benchmark files, reconstructs the normalizer from the training split, and evaluates all 100 test samples to produce per-seed relative error and layer-wise spike rates. This path is used only for model-side characterization.

The \emph{deployment-style} path loads only the checkpoint, precomputed normalization statistics, and a compact 8-sample request bundle drawn from the test set. It excludes one-time bulk dataset loading because re-reading the full benchmark files (approximately 3.2\,GB combined) is not representative of edge inference service and would dominate latency measurements. The timed window therefore captures the request path expected in deployment: preprocessing, model execution, and output decoding.

A parallel \texttt{tegrastats} process samples VDD\_IN power and LPDDR5 RAM every 100\,ms, bracketed by 5\,s idle periods to establish a baseline. Ten warm-up inferences precede 50 timed inferences for latency and 200 back-to-back inferences for the power window. Baseline-subtracted energy per inference is computed by subtracting the idle baseline from the power-window average and dividing by the number of inferences. Seed~20 is additionally profiled with Nsight Systems (CUDA, NVTX, and OS runtime tracing). Because Jetson shares a single LPDDR5 pool between CPU and GPU, we report memory at the system level.

\section{Results}

\subsection{Spike sparsity and deployed cost}

Table~\ref{tab:mainresults} summarizes the main result. On the reference-aligned path, VS-WNO is sparse: mean spike rates decrease from $54.26 \pm 1.65\%$ at the first spiking layer to $18.15 \pm 1.03\%$ at the fourth. On the deployment path, that sparsity does not lower cost. Dense WNO is faster ($53.22 \pm 0.78$\,ms versus $59.55 \pm 0.87$\,ms) and lower-energy ($180.65 \pm 0.69$\,mJ versus $228.04 \pm 2.26$\,mJ), while also achieving lower reference-path error ($1.77 \pm 0.06\%$ versus $1.81 \pm 0.02\%$). On the evaluated Jetson stack, spike sparsity alone is not sufficient; the execution path must also reduce work with spike activity.

The measured request path is the deployment quantity of interest. On that path, sparsity helps only when it reduces executed work. That does not happen here. Although VS-WNO suppresses activation events across layers, it remains slower and higher-energy than the matched dense baseline. On the evaluated Jetson stack, sparsity becomes a deployment benefit only when the execution path makes work scale with spike activity.

\noindent\begin{minipage}{\columnwidth}
\centering
\includegraphics[width=0.85\linewidth]{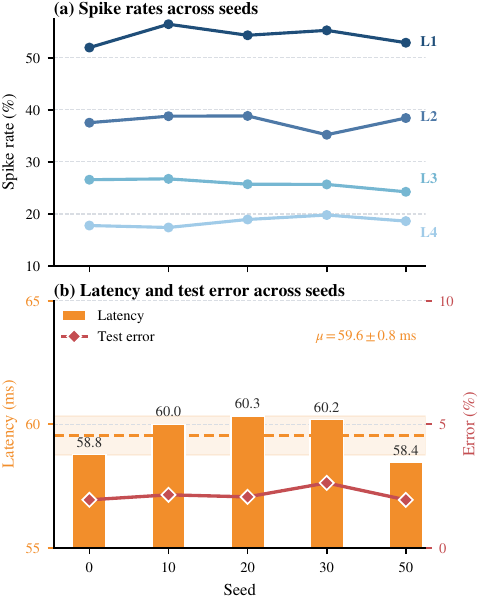}
\captionof{figure}{Per-seed layer-wise spike rates, test error, and warmed latency. Spike rates and error vary across seeds, whereas deployment latency remains stable.}
\Description{Per-seed layer-wise spike rates, test error, and warmed latency for five pretrained VS-WNO checkpoints.}
\label{fig:seeds}
\end{minipage}

\begin{table}[t]
\centering
\caption{Aggregate results across five pretrained seeds $\{0,10,20,30,50\}$ on Darcy rectangular, Jetson Orin Nano 8\,GB.}
\label{tab:mainresults}
\footnotesize
\setlength{\tabcolsep}{4pt}
\renewcommand{\arraystretch}{0.96}
\begin{tabular}{@{}llll@{}}
\toprule
Path & Metric & VS-WNO & WNO \\
\midrule
Ref. & Test error (\%) & 1.81 $\pm$ 0.02 & 1.77 $\pm$ 0.06 \\
Ref. & Spike rate L1 / L2 (\%) & 54.26 / 37.46 & -- \\
Ref. & Spike rate L3 / L4 (\%) & 25.58 / 18.15 & -- \\
Deploy. & Sanity error (\%) & 2.15 $\pm$ 0.28 & 2.28 $\pm$ 0.40 \\
Deploy. & Latency (ms) & 59.55 $\pm$ 0.87 & 53.22 $\pm$ 0.78 \\
Deploy. & Latency p95 (ms) & 60.46 $\pm$ 0.92 & 53.82 $\pm$ 1.19 \\
Deploy. & Dynamic energy / inf. (mJ) & 228.04 $\pm$ 2.26 & 180.65 $\pm$ 0.69 \\
Deploy. & Avg.\ power (W) & 11.34 $\pm$ 0.04 & 10.90 $\pm$ 0.02 \\
Deploy. & GPU peak reserved (MB) & 108.0 & 68.0 \\
Deploy. & Session RAM peak (MB) & 3060.6 $\pm$ 5.1 & 2379.6 $\pm$ 5.9 \\
\bottomrule
\end{tabular}
\end{table}

\FloatBarrier
\subsection{Dense, launch-dominated execution}

Nsight Systems identifies the dominant execution pattern on the deployment path. For VS-WNO, the \texttt{latency\_window} occupies 52.1\% of traced wall time; \texttt{warmup}, \texttt{model\_init}, \texttt{ckpt\_load}, and \texttt{data\_load} account for 31.0\%, 9.3\%, 4.2\%, and 3.5\%, respectively. Within the latency window, \texttt{cudaLaunchKernel} alone accounts for 81.6\% of CUDA API time, with \texttt{cudaMemcpyAsync} at 9.0\% and \texttt{cudaStreamSynchronize} at 3.1\%. GPU execution is likewise dominated by dense operators: grouped/direct convolution contributes 33.3\% of kernel time and depthwise convolution contributes 20.5\%, for a combined 53.8\%.

Dense WNO shows the same pattern: its \texttt{latency\_window} occupies 51.2\% of traced wall time, \texttt{cudaLaunchKernel} accounts for 80.8\% of CUDA API time, and dense convolution kernels account for 62.5\% of GPU kernel time. Thus, both request paths remain dense on the evaluated stack even though only one model is algorithmically sparse.

The mechanism is visible in the implementation. Each wavelet block reconstructs DWT and IDWT modules and moves their filters onto the device inside the forward path. In \texttt{pytorch\_wavelets}~\cite{cotter2019wavelets}, these transforms are ultimately realized through dense convolution operators. Each inference therefore pays dense launch and setup costs regardless of spike count. For sparsity to translate into deployed savings, launched work, arithmetic work, or memory traffic must decrease with spike activity. On the measured Jetson/PyTorch/CUDA path, the Nsight traces indicate that this contraction does not occur. We therefore interpret this result as a property of the evaluated deployment stack and implementation path, rather than as a claim that spiking sparsity is intrinsically ineffective on all substrates.

\begin{figure}[tb]
\centering
\includegraphics[width=0.73\linewidth]{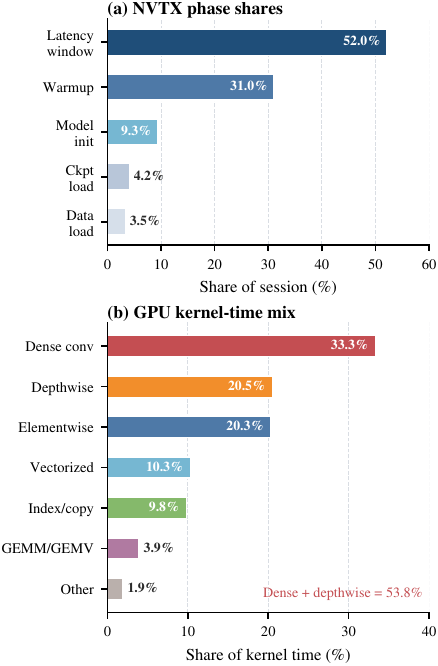}
\caption{Nsight Systems breakdown for VS-WNO seed~20 on the deployment path. (a) NVTX phase shares over the traced session. (b) GPU kernel-time composition within the latency window.}
\Description{Nsight Systems breakdown for VS-WNO seed 20, showing NVTX phase shares and GPU kernel-time composition within the latency window.}
\label{fig:nvtx}
\end{figure}

\FloatBarrier
\subsection{System RAM versus CUDA-reserved memory}

For VS-WNO, CUDA reports a peak reserved footprint of 108\,MB across seeds. That number is misleading on Jetson. \texttt{tegrastats} records deployment-session RAM peaks of $3,060.6$\,MB, reflecting the full unified-memory footprint of Python, PyTorch, the OS kernel, and runtime buffers. The dense WNO baseline shows the same qualitative pattern: CUDA peak reserved is only 68\,MB, while deployment-session RAM peaks at $2379.6$\,MB. The gap is therefore not marginal; for VS-WNO it is roughly $22\times$. On this platform, framework-reported CUDA allocator statistics capture only a small slice of the deployed system state. For capacity planning, system RAM is the relevant metric.

\section{Discussion}

On Jetson Orin Nano, VS-WNO runs correctly under an unmodified PyTorch/CUDA workflow, but its measured sparsity does not reduce deployed latency or energy. The key distinction is between model-side sparsity and system-side work reduction: the former is observable in the activations, whereas the latter requires an execution path whose launched work and data movement contract with spike activity. Our Nsight traces indicate that this condition is not met on the evaluated stack.

For spiking systems, the same measured sparsity can mean different things on different substrates. On commodity GPU stacks it may not change executed work; on event-driven or sparsity-aware substrates it can translate into skipped work and lower energy.

The five-seed VS-WNO reference-path error, $1.81 \pm 0.02\%$, matches the value reported in prior work~\cite{garg2024vswno}, so our claim is about deployment, not model quality. The study is limited to one PDE benchmark, batch size one, PyTorch eager mode, one dense baseline, and one Jetson-class platform. Backends such as \texttt{torch.compile}, TensorRT, or ONNX Runtime could change the kernel mix, and larger batches could amortize launch overhead. Energy is estimated at board level using baseline-subtracted \texttt{tegrastats}. A natural next step is to test sparsity-aware compiled kernels and neuromorphic targets such as Loihi~2~\cite{davies2021advancing}.

\section{Conclusion}

We study five pretrained VS-WNO checkpoints and five matched pretrained dense WNO checkpoints on a Jetson Orin Nano 8\,GB, separating reference-path characterization from deployment-path measurement. VS-WNO exhibits substantial spike sparsity, but that sparsity does not translate into lower deployed latency or energy on the evaluated Jetson stack. Compared with dense WNO, VS-WNO is slower, higher-energy, and slightly less accurate on the measured checkpoints. Nsight Systems shows that the request path remains launch-dominated and dense. For spiking neural operators, deployed efficiency requires more than sparse activations: the execution stack must also make work scale with spike activity.

\bibliographystyle{ACM-Reference-Format}
\bibliography{references}

\end{document}